\documentclass[twoside,twocolumn,9pt]{article}
\usepackage{extsizes}
\usepackage[super,sort&compress,comma]{natbib} 
\usepackage[version=3]{mhchem}
\usepackage[left=1.5cm, right=1.5cm, top=1.785cm, bottom=2.0cm]{geometry}
\usepackage{balance}
\usepackage{mathptmx}
\usepackage{graphicx}
\usepackage{sectsty}
\usepackage{graphicx} 
\usepackage{lastpage}
\usepackage{amsfonts,amssymb}
\usepackage[ruled,linesnumbered]{algorithm2e}
\usepackage{algorithmic}
\usepackage[format=plain,justification=justified,singlelinecheck=false,font={stretch=1.125,small,sf},labelfont=bf,labelsep=space]{caption}
\usepackage{float}
\usepackage{fancyhdr}
\usepackage{fnpos}
\usepackage[english]{babel}
\addto{\captionsenglish}{%
  \renewcommand{\refname}{Notes and references}
}
\usepackage{array}
\usepackage{droidsans}
\usepackage{charter}
\usepackage[T1]{fontenc}
\usepackage[usenames,dvipsnames]{xcolor}
\usepackage{setspace}
\usepackage[compact]{titlesec}
\usepackage{hyperref}

\usepackage{epstopdf}

\definecolor{cream}{RGB}{222,217,201}

\begin{document}

\pagestyle{fancy}
\thispagestyle{plain}
\fancypagestyle{plain}{
\renewcommand{\headrulewidth}{0pt}
}

\makeFNbottom
\makeatletter
\renewcommand\LARGE{\@setfontsize\LARGE{15pt}{17}}
\renewcommand\Large{\@setfontsize\Large{12pt}{14}}
\renewcommand\large{\@setfontsize\large{10pt}{12}}
\renewcommand\footnotesize{\@setfontsize\footnotesize{7pt}{10}}
\makeatother

\renewcommand{\thefootnote}{\fnsymbol{footnote}}
\renewcommand\footnoterule{\vspace*{1pt}%
\color{cream}\hrule width 3.5in height 0.4pt \color{black}\vspace*{5pt}} 
\setcounter{secnumdepth}{5}

\makeatletter 
\renewcommand\@biblabel[1]{#1}            
\renewcommand\@makefntext[1]%
{\noindent\makebox[0pt][r]{\@thefnmark\,}#1}
\makeatother 
\renewcommand{\figurename}{\small{Fig.}~}
\sectionfont{\sffamily\Large}
\subsectionfont{\normalsize}
\subsubsectionfont{\bf}
\setstretch{1.125} 
\setlength{\skip\footins}{0.8cm}
\setlength{\footnotesep}{0.25cm}
\setlength{\jot}{10pt}
\titlespacing*{\section}{0pt}{4pt}{4pt}
\titlespacing*{\subsection}{0pt}{15pt}{1pt}

\fancyfoot{}
\fancyfoot[LO,RE]{\vspace{-7.1pt}\includegraphics[height=9pt]{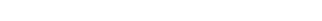}}
\fancyfoot[CO]{\vspace{-7.1pt}\hspace{13.2cm}\includegraphics{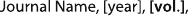}}
\fancyfoot[CE]{\vspace{-7.2pt}\hspace{-14.2cm}\includegraphics{head_foot/RF}}
\fancyfoot[RO]{\footnotesize{\sffamily{1--\pageref{LastPage} ~\textbar  \hspace{2pt}\thepage}}}
\fancyfoot[LE]{\footnotesize{\sffamily{\thepage~\textbar\hspace{3.45cm} 1--\pageref{LastPage}}}}
\fancyhead{}
\renewcommand{\headrulewidth}{0pt} 
\renewcommand{\footrulewidth}{0pt}
\setlength{\arrayrulewidth}{1pt}
\setlength{\columnsep}{6.5mm}
\setlength\bibsep{1pt}

\makeatletter 
\newlength{\figrulesep} 
\setlength{\figrulesep}{0.5\textfloatsep} 

\newcommand{\topfigrule}{\vspace*{-1pt}%
\noindent{\color{cream}\rule[-\figrulesep]{\columnwidth}{1.5pt}} }

\newcommand{\botfigrule}{\vspace*{-2pt}%
\noindent{\color{cream}\rule[\figrulesep]{\columnwidth}{1.5pt}} }

\newcommand{\dblfigrule}{\vspace*{-1pt}%
\noindent{\color{cream}\rule[-\figrulesep]{\textwidth}{1.5pt}} }

\makeatother

\twocolumn[
  \begin{@twocolumnfalse}
{\includegraphics[height=30pt]{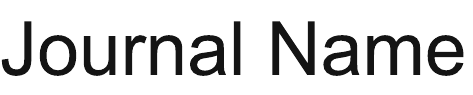}\hfill\raisebox{0pt}[0pt][0pt]{\includegraphics[height=55pt]{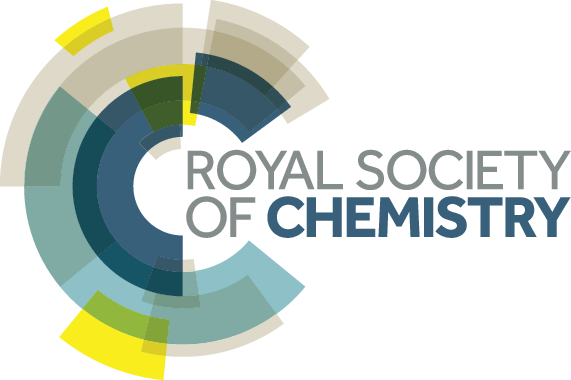}}\\[1ex]
\includegraphics[width=18.5cm]{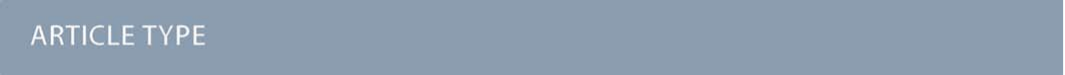}}\par
\vspace{1em}
\sffamily
\begin{tabular}{m{4.5cm} p{13.5cm} }

\includegraphics{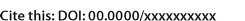} & \noindent\LARGE{\textbf{A high-accuracy multi-model mixing retrosynthetic method}} \\
\vspace{0.3cm} & \vspace{0.3cm} \\

 & \noindent\large{Shang Xiang, Lin Yao, Zhen Wang, Qifan Yu, Wentan Liu, Wentao Guo, and Guolin Ke$^{\ast}$} \\

\includegraphics{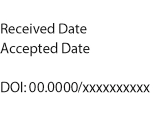} & \noindent\normalsize{The field of computer-aided synthesis planning (CASP) has seen rapid advancements in recent years, achieving significant progress across various algorithmic benchmarks. However, chemists often encounter numerous infeasible reactions when using CASP in practice. This article delves into common errors associated with CASP and introduces a product prediction model aimed at enhancing the accuracy of single-step models. While the product prediction model reduces the number of single-step reactions, it integrates multiple single-step models to maintain the overall reaction count and increase reaction diversity. Based on manual analysis and large-scale testing, the product prediction model, combined with the multi-model ensemble approach, has been proven to offer higher feasibility and greater diversity.} \\

\end{tabular}

 \end{@twocolumnfalse} \vspace{0.6cm}

  ]

\renewcommand*\rmdefault{bch}\normalfont\upshape
\rmfamily
\section*{}
\vspace{-1cm}


\footnotetext{\textit{DP Technology, Beijing, China. E-mail: kegl@dp.tech}}





\section{Introduction}
Chemical synthesis is a fundamental task in drug research. When chemical molecules are either designed or extracted from natural products and require artificial synthesis, chemists devise synthetic routes based on their expertise, utilizing simple and readily available raw materials to synthesize the desired molecules. In 1969, Corey introduced the concept of retrosynthesis\cite{corey1969computer}, a method that begins with the target molecule's structure. By "cutting" a chemical bond, the molecule is broken down into simpler precursor molecules. This process is repeated iteratively until all precursors are readily available.

Retrosynthesis has become a fundamental approach in chemical synthesis. With the rapid advancement of drug design, there is an increasing need to synthesize a vast number of drug molecules for experimental purposes, leading to a heightened demand for molecular synthesis. Traditionally, retrosynthetic analysis methods have relied entirely on chemists, with the design time and quality of synthetic routes dependent on their experience. However, the number of experienced chemists available is insufficient to meet the growing demand. In recent years, Computer-Aided Synthesis Planning (CASP) has emerged, with automated software such as ASKCOS\cite{coley2019robotic} and SYNTHIA\cite{mikulak2020computational} (formerly Chematica\cite{grzybowski2018chematica}) coming to the forefront. Academic research in this field typically divides CASP into two components: single-step reaction generation and multi-step search frameworks, both of which have yielded substantial research outcomes.

Single-step reaction generation is the most fundamental aspect of retrosynthesis, aimed at generating reactions that produce a specified molecule as the product. In Computer-Aided Synthesis Planning (CASP), the template-based method\cite{mikulak2020computational,coley2019robotic,coley2017computer,segler2017neural,dai2019retrosynthesis,fortunato2020data,chen2021deep,yan2022retrocomposer} was the first to be applied, describing chemical reactions through local changes in molecules. This method aligns closely with traditional chemical practices. For instance, Synthia\cite{mikulak2020computational} leverages a vast collection of templates drawn by chemists over decades, resulting in significant achievements. In contrast, Askcos\cite{coley2019robotic} employs algorithms to extract numerous templates and then uses neural networks for intelligent selection. With the advancement of neural networks, template-free\cite{shi2020graph,somnath2021learning,sacha2021molecule,karpov2019transformer,liu2023fusionretro,yao2023node,liu2017retrosynthetic,zheng2019predicting,kim2021valid,tetko2020state,he2022modeling,schwaller2020predicting} methods have also made notable progress. For example, Schwaller encodes\cite{schwaller2020predicting} molecules as strings, views chemical reactions as changes between strings, and applies the transformer's natural language processing (NLP) techniques for reaction generation, achieving groundbreaking results. Additionally, many studies treat molecules as graphs composed of atoms and bonds, with chemical reactions represented as changes in the nodes and edges of the graph. This representation aligns more closely with chemical intuition and has yielded numerous research findings in this field.

The multi-step search framework involves selecting molecules to generate single-step reactions and combining these single-step reactions into synthetic routes. Initially, traditional algorithms such as enumeration and beam search were employed within this framework. In 2018, Segler et al.\cite{segler2018planning} introduced Monte Carlo Tree Search (MCTS), applying the strategy used by AlphaGo to retrosynthesis. Subsequently, the And-Or Tree was found to be highly suitable for retrosynthesis problems, and the corresponding Retro*\cite{chen2020retro} algorithm also performed exceptionally well, becoming the baseline for subsequent retrosynthesis-related research. Additionally, numerous studies based on reinforcement learning\cite{schreck2019learning,kim2021self} and strategies that integrate single-step methods\cite{xie2022retrograph} have been proposed and have achieved notable results.

Due to the scarcity of negative data, it is challenging for the single-step model to assess the feasibility of reactions. Consequently, there is a high likelihood that the generated reactions may not actually occur in reality. In practical applications, most cases still rely heavily on the expertise of chemists. CASP provides several proposed routes, and chemists evaluate the quality of these routes to decide whether to adopt them, subsequently optimizing them based on their own experience. The feasibility and diversity of the routes, which are of primary concern to chemists, are difficult to capture in the current evaluation system. This is not because researchers intentionally overlook this aspect, but rather because these factors are inherently difficult for the algorithm to evaluate and improve upon.

In light of the limited feasibility and diversity of CASP, this paper employs a product prediction model to enhance the feasibility of reactions. It integrates multiple single-step synthesis methods to increase the route completion rate and modifies Retro* to output multiple reaction pathways, all while ensuring route diversity. This approach has led to improved evaluations from chemists for certain actual drug molecules.

\section{Single-step model}
Single-step reaction generation, which involves generating reactions where the product is the target molecule, represents the most fundamental challenge in CASP. While it is possible to retrieve some chemical reactions from existing databases, the sheer number of molecules makes it impractical to have a comprehensive reaction database that can match each query. Without the capability to generate novel reactions, the feasibility of CASP is severely undermined.

Single-step generation methods can typically be categorized into template-based and template-free methods, depending on the approach used to generate the reaction.

\subsection{Template-based method}
Template-based method describes the reaction as a process of transformation of local structure in the molecule, which is similar to the description in chemistry. Divide the reactants into reaction center, the surrounding environment and  the distal part, and do the same process to the products. The reaction can be considered as a process of breaking and connecting bonds in some reaction centers and changing the properties of some atoms, while other parts remain unchanged.

The template method was first utilized in computational calculations and possesses the following characteristics:

Explainability. Templates are comprehensible to chemists, meaning that chemists can understand the reactions represented by the templates. Chemists are able to distinguish between correct and incorrect templates and can also correct erroneous templates based on their expertise.

Reference. Each template has a corresponding real reference reaction. Chemists cannot be familiar with every reaction. When encountering an unfamiliar reaction, finding a reference reaction based on the template can make it easier to judge the feasibility of the reaction or conduct the reaction.

\paragraph*{}
The sources of templates can be roughly divided into two categories: artificial or algorithmic generation.

\subsubsection{Artificial template}
Since the reaction templates are consistent with the reactions typically used by chemists, having chemists write the reaction templates was the earliest method employed. Chemists would convert their knowledge, content from literature, and reactions from databases into templates for single-step reaction generation. Artificial templates have the following advantages:

Accuracy.
Artificial templates conform to chemical reaction mechanisms. Because chemists understand the mechanisms behind reactions, they can accurately determine which parts are essential for the reaction and which parts are irrelevant. Thus, the reaction templates they write are precise, without extraneous parts, and do not omit key components.

Simplicity.
Artificial templates are relatively concise. Each chemical reaction corresponds to at least one reaction template. Even after removing identical templates, there are still millions of different templates in a database with tens of millions of reactions. However, for chemists, the principles they master are far fewer than this quantity. To them, many different templates are based on the same reaction mechanisms. Having more templates is not necessarily better; an excess of templates complicates retrieval and manual verification. Simplified reaction templates are more user-friendly.

Auxiliary information.
Due to the inaccuracy of some reaction templates or other reasons, matching the corresponding template does not necessarily guarantee that the reaction will occur. For instance, other specific functional groups in the reactant might hinder the reaction, or the reaction itself might be highly challenging, only applicable to a limited number of substrates. In such cases, chemists can manually add auxiliary information to these reaction templates to ensure the reaction proceeds smoothly.
  
At the same time, artificial templates also have many disadvantages

It is challenging to write rigorous reaction templates. Although reaction templates are similar to the expressions that chemists are accustomed to, there are still many areas that need attention to make them applicable for computers. There is a lot of information that chemists do not need to explicitly state, such as the fact that precursors should not have unstable functional groups under certain reaction conditions. However, for algorithms, this information needs to be annotated one by one. There are also requirements for the properties of functional groups, such as nucleophilicity and steric hindrance, which are not easy to describe using templates.

Writing reaction templates is an extremely labor-intensive task. Creating a rigorous reaction template is very challenging. A reaction that a chemist can describe with a single formula might require multiple templates to represent accurately. Fully converting a reaction formula into reaction templates is also a significant burden for chemists. Additionally, templates exhibit a long-tail effect. Initially, a small number of templates can cover a large portion of common reactions , but as the number of templates increases, each new template covers fewer and fewer reactions. The number of templates required to cover 20\%,40\%,60\%,80\% of reactions does not increase linearly; it is likely to increase exponentially. Over the past few decades, chemists of Synthia have written 80,000 templates, but tens of thousands of new templates are still added each year. The workload of manually writing reaction templates is simply too overwhelming for chemists.
\subsubsection{Algorithmic template}
Since it is difficult and costly to write templates manually, it is natural to consider using algorithms to generate templates instead of relying on chemists\cite{rdkit,rdchiral}. To generate templates, the algorithm first maps the atoms between reactants and products, identifies the differences between them, and uses the changes in their atoms and bonds as the basis for the templates. This approach allows for the generation of corresponding templates for all reactions, thereby improving the efficiency of template generation. Compared to manually written templates, algorithm-generated templates offer the following advantages:

Cost-effective. It does not require a large number of chemists for mechanical work, and the computational power needed is not high. Even for large volumes of data in databases, conversions can be completed in bulk within a very short period of time.
  
Complete conversion. Algorithm-generated templates can cover all reactions without any omissions due to workload issues.
  
Reference reactions. Since the templates are derived from reaction databases, chemists can directly access reference reactions if they need to learn more about the reactions generated by the templates.

Algorithm-generated templates also have the following disadvantages:

Atom-mapping dependency. Template generation depends on atomic mapping, but currently, the accuracy of atomic mapping cannot reach 100\%, especially for complex reactions. If the atomic mapping is incorrect, the generated template will be unusable or may even produce an erroneous template.

Inaccuracy. The algorithm is not accurate enough in extracting reactions. Even with correct mapping, the reaction center, where the atoms or bonds change, can be well identified. However, many surrounding environments are also necessary for the reaction to proceed. The algorithm cannot identify the size of the required environment as accurately as a human can.
  
Redundancy. The number of templates automatically extracted by the algorithm is too large. Enumerating millions of templates can result in an unmanageable computational load. Additionally, without limitations, the number of generated reactions will be difficult to control. In many instances, templates need to be clustered, but clustering is not a simple task.
  
Infeasibility. Templates extracted by the algorithm have not been manually verified, lack corresponding auxiliary information, and are too numerous to be used as easily as manually created templates. To compensate for this shortcoming, there are generally two methods: 1. Use reaction similarity: Find reactions in the database whose products are similar to the target molecule, and the generated reaction will be expected to be similar to the database reaction. 2. Use neural networks to intelligently select templates based on the characteristics of the target molecule.
  
\subsection{Template-free method}
Due to the inherent challenges templates face in accurately and comprehensively describing chemical reactions, template-free methods have increasingly gained prominence, propelled by advancements in neural networks. These template-free approaches can be broadly categorized into two types based on how the reaction molecules are represented: SMILES-based methods\cite{liu2017retrosynthetic,zheng2019predicting,kim2021valid,tetko2020state,he2022modeling,zhong2022root,lin2020automatic} and graph-based methods\cite{zhong2024recent,shi2020graph,chen2023g,zhong2023retrosynthesis,sacha2021molecule}.

\subsubsection{SMILES-based method}
In computational chemistry, molecules can be characterized using one-dimensional or two-dimensional methods. The one-dimensional method involves encoding the molecule, such as the SMILES, which represents molecules using strings based on a traversal sequence. This format has become the most widely used molecular representation in algorithms. SMILES and molecular structures can be easily interconverted, with the conversion from SMILES to molecules being an injective function. However, the same molecule can have multiple different SMILES representations. The introduction of canonical SMILES ensures that each molecule corresponds uniquely to a canonical SMILES string. Based on this, chemical reactions can be viewed as transformations from one string to another. 
Thanks to the rapid advancements in natural language processing (NLP), the ability of models to understand text has significantly improved. In particular, Transformer models have been used for single-step reaction generation and have achieved excellent results.

SMILES-based model has following advantages:

Natural Language Processing (NLP) has been extensively researched, with model updates and iterations happening at a rapid pace, making the functionality extremely powerful. Ready-made models can be used, and the training process is straightforward.

This model is easy to apply, utilizing SMILES notation to represent molecules from beginning to the end, without the need for additional processing of inputs and outputs.

SMILES-based model also has following disadvantages:

SMILES are artificially encoded representations of molecules, and the model first needs to understand the encoding principles, which increases the difficulty of comprehension for the model. Additionally, the intrinsic information of molecules differs from natural language processing; models that perform well in NLP may not necessarily yield the same effectiveness in reaction generation.

The strings output by the model are not always valid, and many outputs are different SMILES forms of the same molecule. Although with the development of models, most of the output strings are valid SMILES forms, there are still some that are invalid. While the canonical SMILES format of a molecule is unique, it is very difficult for the model to learn its rules.

\subsubsection{Graph-based method}

In chemistry, skeletal formula is often used to represent molecular structures, where atoms are depicted as nodes and chemical bonds as edges. This is a two-dimensional representation of molecules. Through skeletal formula, chemist can intuitively see the composition of the molecule, the connectivity between atoms, and the symmetry of the molecule, which is crucial for predicting the physical and chemical properties of the molecule. The skeletal formula is fundamentally similar to graph structures in data structures. In computers, the coordinates of each atom are ignored, and only the chemical bond relationships between atoms are preserved.

In the representation of graph structures, chemical reactions are viewed as a series of transformations on the graph, including the addition, removal, and attribute changes of atoms and chemical bonds. Neural network models generate molecular graphs based on molecular structures, output transformation sequences, and then generate reactants based on these sequences.

Graph-based model has following advantages:

The graph-based model fully exploits the molecular topological structure information and the alignment information of product atoms, making the data representation consistent with its chemical nature. 

The graph corresponds one-to-one with the molecular structure, and by simply restricting the number of bonds per atom, valid molecules can be obtained. 

Graph-based model also has following disadvantages:

The order of graph editing is not unique, requiring the setting of complex artificial rules; otherwise, a large number of identical results may be generated. 

There are no effective data augmentation methods, leading to high data requirements. 

From molecule to graph and back to molecule, there is currently no complete serialization and deserialization process. In terms of spatial structure, chirality, as 3D spatial information, suffers from information loss when mapped to a 2D structure.

\section{Infeasible reaction analysis}
Our expectation for the single-step model is that it should be capable of providing all synthetically feasible reactions that can synthesize the target molecule. By "synthetically feasible", we mean that the reactants should be simpler and easier to synthesize compared to the product. Ideally, if we had a dataset containing all possible reactions for synthesizing each molecule, two metrics—
\begin{equation*}
\begin{aligned}
&recall = \frac{|\{synthesis\ reactions\}\cap\{output\ reactions\}|}{|\{synthesis\ reactions\}|}\\
&precision = \frac{|\{synthesis\ reactions\}\cap\{output\ reactions\}|}{|output\ reactions|}
\end{aligned}
\end{equation*}
where $synthesis\ reactions, output\ reactions$ 
means reactions for synthesizing target molecules in the database and model output reactions respectively—could effectively reflect the model's capability. However, existing datasets typically consist of collections of several reactions without summarizing synthesis methods for individual molecules, and most molecules have only one synthesis method. Under these circumstances, using recall and precision metrics cannot accurately assess the model's performance. Recall is generally replaced by the top-N hit rate, which measures whether the required reaction is included in the top N reactions suggested by the model, reflecting the model's learning ability. However, precision lacks a suitable alternative metric to constrain the model's output and prevent incorrect reactions. Therefore, even if the top-N hit rate reaches 100\%, it does not guarantee that all suggested reactions are feasible; it only ensures that a correct reaction is among them. This can result in many infeasible reactions being mixed with the correct ones, making it impossible to differentiate between them and rendering the generated synthesis routes impractical in real-world applications.

Chemists' ideal expectation for CASP is to provide the "optimal" synthesis route. However, this is quite challenging and beyond the capabilities of current algorithms. As a compromise, chemists hope that CASP can offer multiple "semi-finished routes", where some reactions may be infeasible but the overall strategy is sound, allowing for manual corrections to make the reactions usable. This approach can still save chemists a significant amount of work. Since the "semi-finished routes" provided by CASP do not always meet chemists' needs, there is also a desire for a variety of routes to choose from. However, current CASP systems often fall short of these expectations, primarily due to two reasons: insufficient reaction feasibility, with many generated reactions being impossible in the real world, and a lack of diversity in the suggested reactions.

Reaction feasibility is crucial for retrosynthetic analysis, but there is currently no suitable algorithm to assess it, and the databases lack negative data for infeasible reactions. As a result, existing methods struggle to avoid generating infeasible reactions, which can be categorized into the following types:

\subsection{Invalid molecule}
This error is most common in SMILES-based models, because the model does not have strong constraints to ensure that the output string conforms to the SMILES format. Although the current SOTA SMILES-based model has a very high accuracy rate, there are still some syntax errors. Invalid molecules in the graph base model generally violate some chemical rules, such as incorrect atomic valence bonds and incorrect positions of aromatic carbons. The incorrect molecules generated by the template base method are mostly due to template writing errors. Such problems can be easily identified, and will not affect the accuracy of the routes, but only the amount of reaction generated.
\subsection{Incomprehensible reaction}
This is a problem that may occur with template-free methods. These black-box methods are not interpretable. Some very peculiar changes, such as a sudden alteration of an atom in a molecule or the unknown origin of certain structures, have become less frequent as model performance improves. However, many changes that are difficult to understand from a chemical perspective still occur in template-free methods. Template-based methods, on the other hand, rarely have such issues. The template itself represents the chemical process, allowing chemists to easily understand how the reaction is generated. If the template does not conform to chemical principles, it can be directly discarded.

\begin{figure*}[htbp]
\centering
\includegraphics{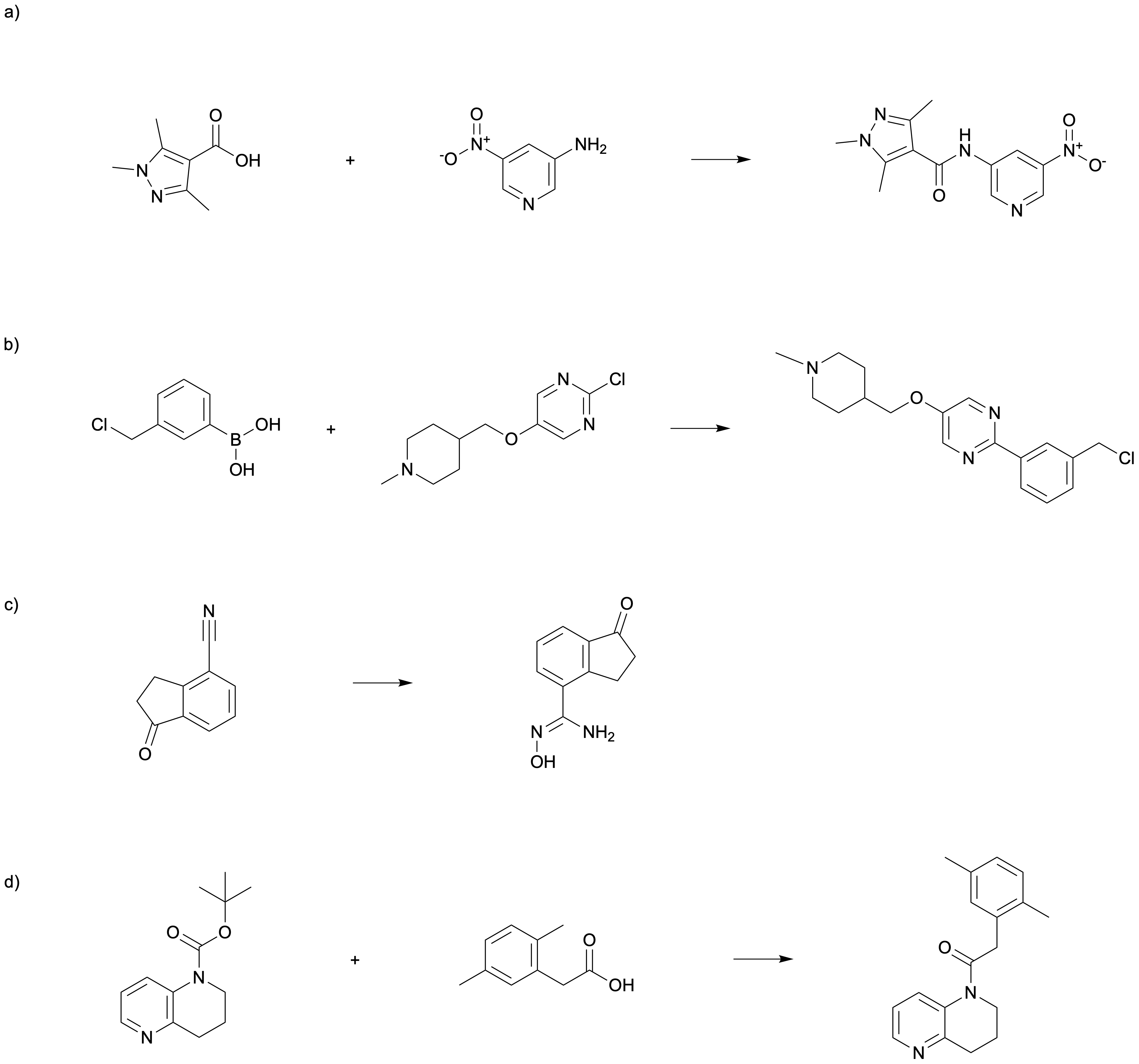}
\caption{Infeasible reactions examples: 
(a)A Buchwald-Hartwig coupling reaction, but the presence of functional groups that hydrolyze under alkaline conditions will affect the reaction (such as carboxylic acid esters). (b)A Suzuki coupling reaction, but under this condition, benzyl chloride cannot remain stable and will also be reacted. (c)It is not feasible to do addition reaction for cyano groups without protecting the ketone carbonyl group, but simply adding a protecting group to the cyano group makes it feasible. (d)This is a two-step reaction of debocification and acid amine addition, which cannot be achieved in a single step reaction.}
\end{figure*}

\subsection{Insufficient reactivity}
Chemists can understand these and subsequent reactions, and the reactions can be described by templates. Such errors typically arise when the reaction center or its surrounding area does not satisfy the necessary requirements for the reaction. In the case of template methods, this is often due to imprecise template definitions and the absence of constraints, which results in the inclusion of non-reactive scenarios. For template-free methods, a lack of negative data can lead to underfitting of the model.
\subsection{Other competing functional groups}

  Under the corresponding reaction conditions, the reaction center satisfy the reaction requirements, but the competitive functional group outside the reaction center will react first resulting in reaction failure. This situation is very common. In artificial templates, experts will write down mutually exclusive functional groups in the corresponding template, because these external functional groups will affect the reaction.
  
\subsection{Need to add protecting groups}
  
  It can be seen as a special case of d, but the competitive functional group can be easily protected before the reaction and deprotected after the reaction to avoid reaction failure. It just makes the actual reaction route longer than the expected route. The route can be used after minor modifications and is often not considered a failed route.
\subsection{Multi-step reaction}
  Due to the irregular collection of reactions in the database, multi-step common reactions are sometimes shortened to one-step reactions, and the same problem occurs when training a single-step model. Chemists consider that such reactions are not feasible, but in actual applications, they can be simply split into two steps and are not considered to be incorrect reactions.

\section{Optimized for chemist friendliness}
\subsection{Product prediction model}
To verify the feasibility of a single-step model using computational chemistry methods is essentially impractical due to the lack of necessary reaction conditions and the high computational costs involved. In the field of retrosynthesis, a product prediction model can assist in evaluating the likelihood of a reaction occurring. After generating a single-step reaction model, the predicted reactants are evaluated using the product prediction model to determine whether the predicted product matches the target molecule. If they match, the output reaction is considered feasible. Pre-trained models capitalize on self-supervised learning methodologies to harness the wealth of unlabeled data for knowledge distillation, offering significant utility in domains such as bioinformatics where data scarcity is prevalent. Given the well-developed pre-training of molecules based on SMILES, we here consider product prediction as a translation task between the SMILES of reactants and products. 

BARTSmiles\cite{chilingaryan2022bartsmiles}, a specialized variant of the BART (Bidirectional Auto-Regressive Transformers)\cite{lewis2019bart}model, is a generative masked language model designed for molecular representation. Notably, BARTSmiles stands out for its extensive pre-training on a comprehensive dataset of over 1.7 billion molecules, coupled with a tailored pre-training strategy that is specifically optimized for the chemical domain, thereby delivering exceptional capabilities and broad applicability in chemical research.
We fine-tune the BARTSmiles model on reaction datasets. The model utilizes a bidirectional encoder $f(\theta_{en})$ to process the input SMILES strings of reactants, generating a sequence of hidden states. These hidden states are then fed into the decoder $g(\theta_{de})$ to predict the product molecules. 

To facilitate more comprehensive training on reaction datasets, we have employed the Byte Pair Encoding (BPE) algorithm to extend the BARTSmiles vocabulary, thereby encompassing a broader range of elemental tokens. Furthermore, by utilizing atom mapping between reactants and products, we have implemented various data augmentation strategies to expand the diversity of the reaction dataset. The product prediction model trained in this manner can effectively identify various patterns within the reactants. By utilizing decoding algorithms such as beam search, it is able to generate diverse product prediction outcomes, which serve as a means to assess the effectiveness of reaction generation models.

By using the product prediction model, we can effectively determine whether the reactants predicted by the single-step model can synthesize the target molecule, thereby filtering out infeasible reactions.

\begin{figure*}[htbp]
\centering
\begin{minipage}{0.45\linewidth}(a)
\centering
\includegraphics[width=1\linewidth]{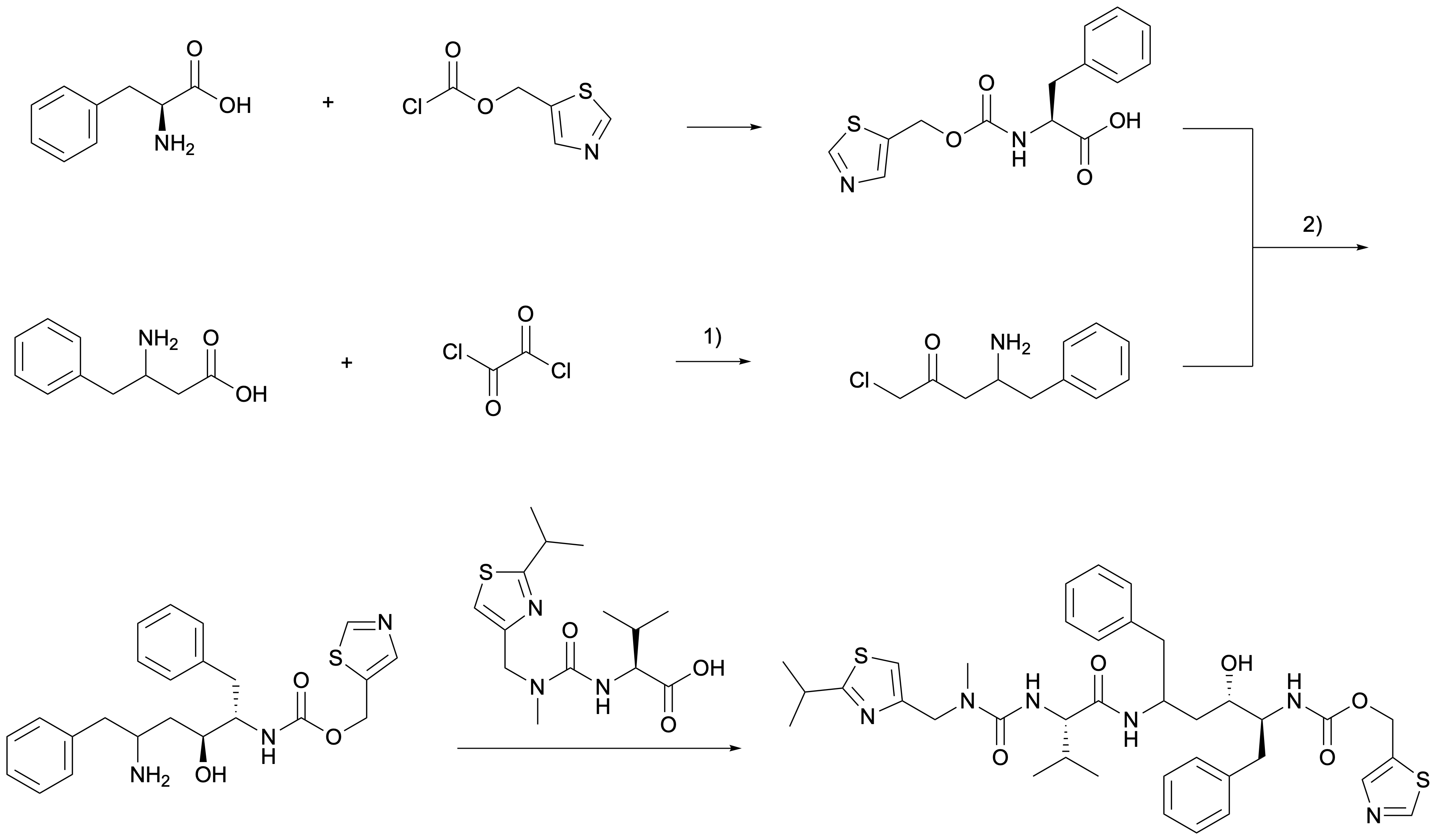}
\label{fig:4-1}
\end{minipage}
\hfill
\begin{minipage}{0.45\linewidth}(b)
\centering
\includegraphics[width=1\linewidth]{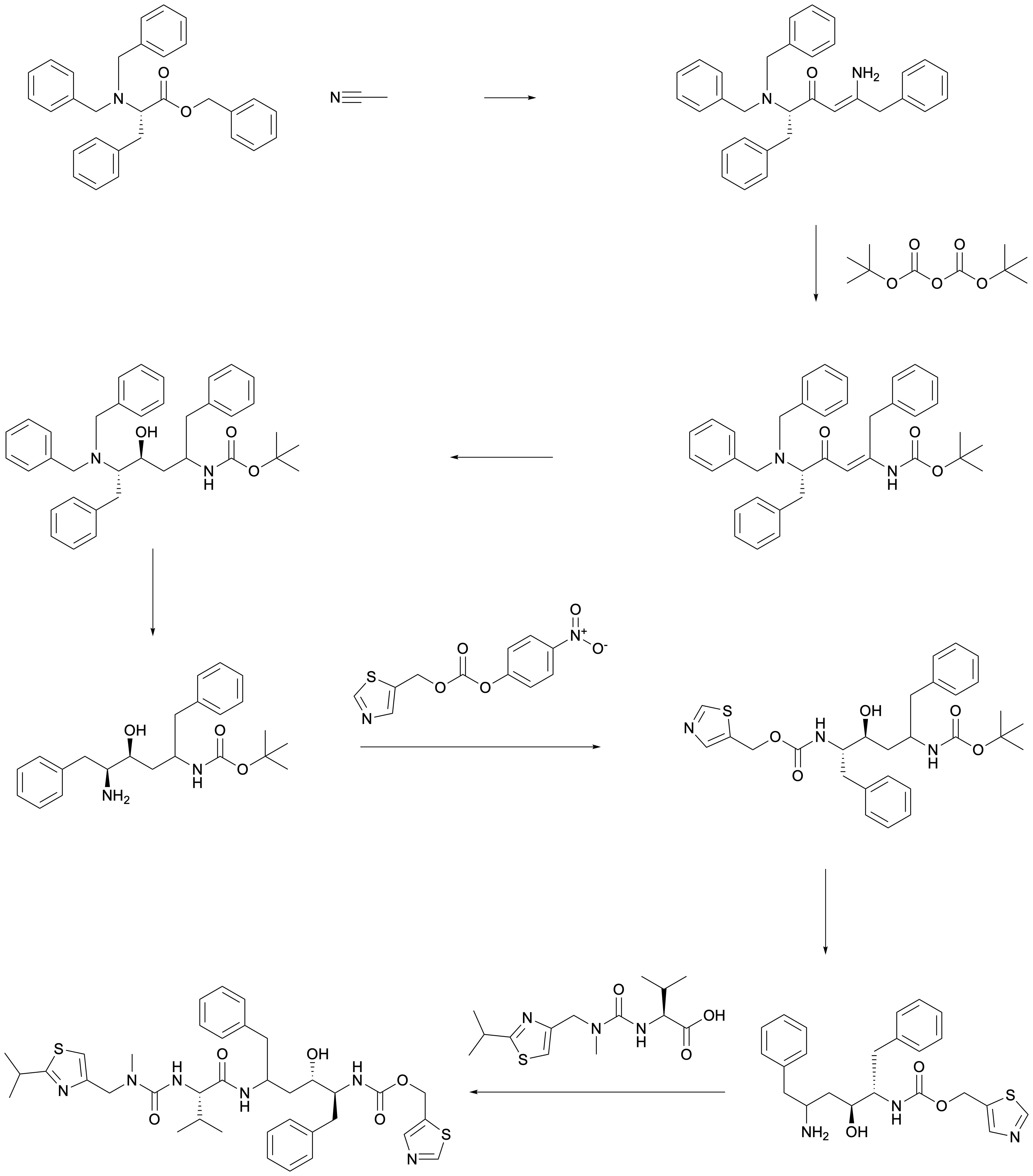}
\label{fig:3-2}
\end{minipage}
\quad
\caption{Two algorithm-generated synthetic routes for nirmatrelvir:
(a) Based on a similarity template model without product prediction model filtering. The route is short, but existing two reactions which are difficult to achieve. In reaction 1, the product should be an acyl chloride, not an $\alpha$-chloro ketone. Additionally, the exposed amino group may also affect the reaction. In reaction 2, the functional groups, oxidation states, and atom counts of the reactants and products do not correspond. If the two reagents were mixed, it is highly likely that the carboxylic acid would replace the Cl.
(b) Based on a similarity template model with product prediction model filtering. The route is longer, but the reactions are comparatively easier to achieve.
}
\end{figure*}

\subsection{Reaction generate model mixed}
However, utilizing the product prediction model presents a significant challenge: it reduces the number of available reactions and significantly lowers the route completion rate. Additionally, some feasible reactions may be erroneously excluded, leading to the elimination of originally correct routes. To address this issue, we can enhance the performance of the product prediction model and increase the number of single-step reactions. By integrating various single-step models, we can provide a broader range of single-step results. To ensure greater diversity in these outcomes, different types of reactions can be retained after filtering through the product prediction model. We have selected three distinct single-step models: the template method based on similarity, the BARTSmiles\cite{chilingaryan2022bartsmiles} model based on SMILES, and the NAG2G\cite{yao2023node} model based on graphs.

The template method based on similarity involves identifying reactions in a database whose products resemble the target molecule and using templates derived from these reactions to generate corresponding reactions. This approach adheres to chemical logic: if a similar molecule can react, we expect the target molecule to undergo a comparable reaction. Chemists often utilize this logic to design synthetic routes. The similarity is measured using the ECFP distance, a method of distance calculation that is applied across many fields. Generally, molecules with similar chemical structures have relatively close ECFP distances. However, when the structures are not very similar, chemists focus more on the chemical properties, and the effectiveness of the ECFP method diminishes.

The BARTSmiles\cite{chilingaryan2022bartsmiles} model is used to transform the SMILES strings of target molecules to generate reactants. Given the efficiency and standardization of SMILES as a method for encoding molecular structure information, we employed a reaction prediction model based on SMILES. We fine-tuned BARTSmiles\cite{chilingaryan2022bartsmiles} on a reaction dataset. Similar to the approach used in product prediction models, but with the input and output reversed, we used the product as the input and the reactants as the output. We continued to adopt a data augmentation strategy based on atom mapping and utilized an extended vocabulary. It is worth noting that the main purpose of the reaction generation model is to obtain more diverse synthetic routes related to the target molecule. Therefore, adjusting parameters such as temperature and length penalty in beam search can more effectively enhance the diversity of candidate reactions.

The NAG2G\cite{yao2023node} model is a model that obtains reactants through the generation of graph editing sequences. By using graph structures to represent molecules, it effectively leverages the topological relationships between reactants and products, transforming the reaction into several operations on the molecular graph. NAG2G utilizes 2D molecular graphs and conformations to retain comprehensive molecular details and incorporates product-reactant atom mapping through node alignment, which determines the order of the node-by-node graph output process in an auto-regressive manner. Through rigorous benchmarking and detailed case studies, NAG2G has demonstrated excellent 

We anticipate that, given the abundance of data, template-free models can effectively learn common chemical reactions. However, caution is advised when dealing with less common reactions, and it is preferable to select those with similar references available in the database.

\subsection{Route diversity improvement}
We choose Retro* as the multi-step search framework because in the context of multi-model mixing, the additional information (such as similarity, model selection probability) provided by various single-step models is difficult to measure uniformly. Retro* does not require reaction-related information; it only needs a molecular valuation model, $V(M)\rightarrow\mathbb{R}$, to predict the difficulty of synthesizing the molecule, providing a reference for the molecular expansion sequence in Retro*.

In Retro*, each molecule maintains only one current "optimal" synthesis method. Additionally, when $V(M) \ge V_T(M)$, which $V_T(M)$ means the true synthesis difficulty of molecule M, it can be guaranteed that the found route is the "optimal" route. However, in practical applications, the reactions generated by the algorithm are not always feasible, and its "optimal" route may not be better than a non-optimal route. Therefore, we need to generate multiple good synthesis routes for the synthesizer to choose from.

We extend the optimal route maintained for each molecule in Retro* to include the best k routes. 

\begin{equation}
\begin{split}
&rl_{k}(R|T) = c(R) + \mathop{}\limits_{m_{i}\in{ch(R)}}rl_{k}(m_{1}|T) \bigotimes rl_{k}(m_{2}|T) \bigotimes ...\\
&rl_{k}(m|T) = \left\{
\begin{array}{lr}
  (V_{m}, \inf, \inf, ...), & m\in \mathbb{F}(T) \\
  \mathop{}\limits_{R_{i}\in{ch(m)}}rl_{k}(R_{1}|T) \bigoplus rl_{k}(R_{2}|T) \bigoplus ... , & otherwise \\
  \end{array}
\right.
\end{split}
\end{equation}

which \textit{R}, \textit{m}, \textit{c(R)}, $\mathbb{F}$, $V_m$, \textit{ch} means 
       reaction,   molecule,  reaction cost, leaf node of search tree, molecular valuation, children respectively.
And $(a_{1}, a_{2}, ..., a_{k}) \bigotimes (b_{1}, b_{2}, ..., b_{k}) = sorted(a_{1}+b_{1}, a_{1}+b_{2}, ... ,a_{k}+b_{k})[1:k]$ and $(a_{1}, a_{2}, ..., a_{k}) \bigoplus (b_{1}, b_{2}, ..., b_{k}) = sorted(a_{1}, b_{1}, a_{2}, b_{2}, ... ,a_{k}, b_{k})[1:k]$

Under this definition, non-optimal routes may exhibit cycles, where molecule one generates molecule two, and molecule two, in turn, generates molecule one. We impose restrictions on the routes to prohibit such cycles.

After testing, we found that the algorithm can generate multiple different synthesis routes. However, the generated routes often have high similarity; many times, two routes differ only by a small number of substrates, which, in chemical terms, means they are essentially the same route. To address this issue, we implemented a similarity suppression strategy. 

For multiple similar routes of the same molecule, we add a penalty coefficient to their synthesis difficulty during maintenance, based on the number of repeated reactions in the route compared to other routes. This aims to provide multiple distinct routes, ensuring a diverse set of options.

\begin{algorithm}[htbp]
\caption{Similar routes suppression}
\KwIn{synthetic routes $p_{1}$, ... $p_{n}$, k}
\KwOut{best k routes}

Sort routes by route\_cost;

\For{$i=1;i \le n$} 
{
    $repeat_i = 0;$
    
    \For{$j=1;j \le i$} {
    
        $repeat_i = max(repeat_i, |route_{i} \cap route_{j}|);$
        
    }
    
    $penalty_i = (1 + 0.1*repeat_i)^2;$
    
    $suppression\_cost_i = route\_cost_i * penalty_i;$
    
}

Sort routes by suppression cost;

\Return $p_{1..k}$;
\end{algorithm}

\subsection{Evaluate function}

To improve route search, we retrained the evaluation function model to assess the synthesis cost of molecules. This model generates a predicted value for each molecule, representing the estimated synthesis cost, which can better estimate search routes and select suitable molecules for expansion.

We define the synthesis cost of a route as:
\begin{equation}
     c(Route) = \log(price + 1) + |Route|
\end{equation}
where $price$ represents the estimated synthesis price of the molecule, and $|Route|$ indicates the number of reaction steps.
The synthetic cost of a molecule is the cost of its optimal synthesis route.

We extract synthesis routes from the database and then find the optimal synthesis route for each molecule to obtain the corresponding synthesis cost, which is used as training data for the model.

Finding all synthesis routes is challenging. To better capture the complexity of molecule synthesis, we also adopt a recursive method to generate synthesizable molecules. Starting from the raw material library, we first identify molecules that can be synthesized in one step, then iteratively generate molecules that can be synthesized through more reaction steps, while calculating the estimated synthesis price for each molecule based on the reactions. For each newly generated molecule, its synthesis cost is calculated according to the above formula and used as the predicted value.

Finally, we apply the BARTSmiles\cite{chilingaryan2022bartsmiles} model. Using the SMILES of the molecule as input and the estimated synthesis cost as output, we obtain the evaluation model.

\section{Experiment}
\subsection{Dataset}
We utilized a commercially available reaction dataset, Pistachio \cite{pistachio}, from NextSortware to train our reaction generation model, product prediction model, and evaluate model. The entire dataset comprises approximately 13 million reactions up to 2023, categorized into 10 reaction classes. For multi-product reactions, we divided them into multiple single-product reactions. After removing duplicates and erroneous entries, the remaining ~5 million data points were randomly split into training, validation, and test sets, with proportions of 80\%, 10\%, and 10\%, respectively.

To train our validation model, we constructed validation data using the reaction database and a list of commercially available building blocks from eMolecules, which consists of 300 million building blocks. Given the list of building blocks, we analyzed each molecule that appeared in the Pistachio reaction data to determine if it could be synthesized using existing reactions within the Pistachio dataset. For each synthesizable molecule, we selected the least costly synthesis routes, ensuring the end points were available building blocks from eMolecules. Since there is no price data in eMolecules, we set the building block price to zero when evaluating route costs.

The detailed descriptions of the reaction generation model, product prediction model, and evaluation model are provided in Section 4.

\subsection{Result}
To verify the effectiveness of the product prediction model, human evaluation is required, which prevents the use of large-scale data. We selected 32 small molecule drugs approved by the FDA in 2023 as the test dataset and conducted human evaluations on the routes generated by different single-step methods. Chemists scored each generated chemical reaction step, where 0 points indicate the reaction is infeasible, 1 point indicates the feasibility is uncertain and requires database checks or experiments to determine, and 2 points indicate the reaction is feasible.

\begin{table}[h]
\small
  \caption{Synthesis route for 2023 drug}
  \label{tbl:2023drug}
\begin{tabular*}{0.48\textwidth}{@{\extracolsep{\fill}}llllll}
    \hline
    single-step & checker & 0 points & 1 point & 2 points & average\\
    \hline
    NAG2G & $\times$ & 7.55\% & 16.98\% & 75.47\% & 1.68 \\
    NAG2G & \checkmark & 2.04\% & 12.24\% & 85.71\% & 1.84 \\
    BARTSmiles & $\times$ & 21.95\% & 7.32\% & 70.73\% & 1.49 \\
    BARTSmiles & \checkmark & 16.67\% & 4.76\% & 78.57\% & 1.62 \\
    similarity & $\times$ & 12.00\% & 24.00\% & 64.00\% & 1.52 \\
    similarity & \checkmark & \textbf{0.00}\% & \textbf{11.11}\% & \textbf{88.89}\% & \textbf{1.89} \\
    \hline
  \end{tabular*}
\end{table}

Table~\ref{tbl:2023drug} shows that the reaction quality generated by three different single-step methods varies. Without using a product prediction model for filtering, the NAG2G method had the highest reaction quality, with the most reactions scoring 2 points and the fewest reactions scoring 0 points, resulting in an average point of 1.68. The BARTSmiles method had an average point of 1.49, while the similarity-based template method had an average point of 1.52. However, when a reaction prediction model was incorporated, the reaction quality significantly improved across all methods. The NAG2G method, BARTSmiles method, and similarity-based template method showed increases of 0.157 points, 0.131 points, and 0.368 points, respectively. Among these, the similarity-based template method showed the most significant improvement in reaction quality, with a 12\% reduction in 0-point reactions and a 24.89\% increase in 2-points reactions.

We examined the impact of model mixing on route completion rate and route length under large-scale testing. We selected two test sets, n1 and n5\cite{genheden2022paroutes}, which were artificially constructed for large-scale retrosynthetic testing. Both n1 and n5 contain 10,000 molecules, with different sets of raw materials, and the molecules in n5 are relatively harder to synthesize compared to those in n1.

\begin{table*}
\small
  \caption{Synthesis route for N1}
  \label{tbl:n1}
\begin{tabular*}{\textwidth}{@{\extracolsep{\fill}}lllll}
    \hline
    single-step model & checker & success rate & average length of top1 path & average length of top10 path \\
    \hline
    NAG2G & $\times$ & \textbf{99.81}\% & \textbf{2.23} & \textbf{3.01} \\
    BARTSmiles & $\times$ & 99.48\% & 2.55 & 3.74 \\
    similarity & $\times$ & 96.68\% & 2.81 & 3.70 \\
    NAG2G & \checkmark & 83.04\% & 2.91 & 4.14 \\
    BARTSmiles & \checkmark & 78.37\% & 3.16 & 4.72 \\
    similarity & \checkmark & 73.37\% & 3.37 & 4.68 \\ 
    NAG2G+BARTSmiles & \checkmark & 85.15\% & 2.82 & 3.93 \\
    NAG2G+similarity & \checkmark & 86.81\% & 2.78 & 3.78 \\
    BARTSmiles+similarity & \checkmark & 86.04\% & 2.94 & 4.08 \\
    NAG2G+BARTSmiles+similarity & \checkmark & \textbf{87.43}\% & \textbf{2.71} & \textbf{3.66} \\
    \hline
  \end{tabular*}
\end{table*}

\begin{table*}
\small
  \caption{Synthesis route for N5}
  \label{tbl:n5}
\begin{tabular*}{\textwidth}{@{\extracolsep{\fill}}lllll}
    \hline
    single-step model & validation model & success rate & average length of top1 path & average length of top10 path \\
    \hline
    NAG2G & $\times$ & \textbf{99.75}\% & \textbf{2.66} & \textbf{3.36} \\
    BARTSmiles & $\times$ & 98.83\% & 2.99 & 4.04 \\
    similarity & $\times$ & 95.66\% & 3.23 & 4.00 \\
    NAG2G & \checkmark & 83.39\% & 3.35 & 4.43 \\ 
    BARTSmiles & \checkmark & 77.99\% & 3.66 & 5.05 \\
    similarity & \checkmark & 72.18\% & 3.83 & 4.94 \\ 
    NAG2G+BARTSmiles & \checkmark & 85.45\% & 3.24 & 4.21 \\
    NAG2G+similarity & \checkmark & 87.00\% & 3.15 & 4.05 \\
    BARTSmiles+similarity & \checkmark & 86.39\% & 3.38 & 4.36 \\
    NAG2G+BARTSmiles+similarity & \checkmark & \textbf{87.63}\% & \textbf{3.09} & \textbf{3.94} \\
    \hline
  \end{tabular*}
\end{table*}

Table~\ref{tbl:n1} presents a quantitative analysis of the impact of various strategies on the model's performance on the N1 dataset, with a particular focus on route completion rates. When no product prediction model is used for filtering, the NAG2G achieves a route completion rate of 99.81\%, with an average optimal route length of 2.23 and an average top 10 route length of 3.01. These metrics significantly outperform those of BARTSmiles and similarity. The route completion rates for BARTSmiles and similarity are also high, at 99.48\% and 96.68\% respectively. However, after incorporating the product prediction model, all three models show a decline in these metrics. When different models are combined, compared to single models, the completion rates increase and the average route lengths decrease. Notably, even the best-performing single model, NAG2G, shows considerable improvement when combined with the relatively worst-performing similarity model. When all three models are combined, the metrics improve further, with a route completion rate of 87.43\%, representing a 4.39\% increase over the best single model NAG2G, which achieved 83.04\%. Nevertheless, this is still significantly lower than the completion rate without the product prediction model filtering.

Table~\ref{tbl:n5} details the performance of each strategy on the N5 dataset, and the trends observed are consistent with those in Table~\ref{tbl:n1}.

To validate the effectiveness of the route similarity suppression strategy, we defined the  $repetition\ rate = \frac{|reactions|}{|unique \ reactions|} - 1$. On the n5 dataset, we analyzed the top 10 output routes. We found that the similarity suppression strategy effectively increased the diversity of the routes.

Table~\ref{tbl:repetition rate} shows the impact of applying similarity suppression strategies. Under the original Retro* framework, the repetition rates of the top 10 routes generated by the reaction generation models NAG2G and BARTSmiles were 78.67\% and 86.19\%, respectively. After adding the similarity suppression strategy, the repetition rates of the top 10 routes generated by NAG2G and BARTSmiles decreased to 63.44\% and 71.23\%, respectively.

\begin{table}[h]
\small
  \caption{Top 10 routes reaction repetition rate of N5}
  \label{tbl:repetition rate}
\begin{tabular*}{0.48\textwidth}{@{\extracolsep{\fill}}lll}
    \hline
    single-step model & origin Retro* & similarity suppression \\
    \hline
    NAG2G & 78.67\% & \textbf{63.44}\% \\
    BARTSmiles & 86.19\% & 71.23\% \\
    \hline
  \end{tabular*}
\end{table}

\section{Conclusions}
Utilizing product prediction models to filter reactions can significantly improve reaction feasibility, although it also reduces the number of modeled reactions and lowers the route completion rate. To offset this decline, a combination of various single-step reaction models can be employed, which boosts the route completion rate, though it still remains lower than it was before filtering. This reduction could be due to the product prediction model filtering out some critical synthesis reactions. Moreover, implementing similarity inhibition strategies can effectively enhance route diversity, offering chemists a broader range of options.

\section*{Author contributions}
Shang Xiang designed, conducted the research, and wrote the manuscript. Lin Yao is responsible for the implementation and application of NAG2G. Zhen Wang is responsible for the SMILES-based model. Qifan Yu designed the value function and train the evaluate model. Wentan Liu provided chemical support for the project and scored the route. All authors read and approved the mannal manuscript.

\section*{Conflicts of interest} 
There are no conflicts to declare.

\section*{Data availability}
Data available on request from the authors.



\balance

\renewcommand\refname{References}

\bibliography{rsc} 
\bibliographystyle{rsc} 
\end{document}